\documentclass[a4paper,twoside]{article}

\usepackage{epsfig}
\usepackage{subcaption}
\usepackage{calc}
\usepackage{amssymb}
\usepackage{amstext}
\usepackage{amsmath}
\usepackage{amsthm}
\usepackage{multicol}
\usepackage{pslatex}
\usepackage{apalike}
\usepackage{algorithm2e}
\usepackage[bottom]{footmisc}
\usepackage{SCITEPRESS}     

\begin{document}

\title{Decomposer: Semi-supervised Learning of Image Restoration and Image Decomposition }


\author{\authorname{Boris Meinardus\sup{1}, Mariusz Trzeciakiewicz\sup{1}, Tim Herzig\sup{1}, \\Monika Kwiatkowski\sup{2}, Simon Matern\sup{2}, Olaf Hellwich\sup{2}}
\affiliation{\sup{1}Technische Universität Berlin, Germany}
\affiliation{\sup{2}Computer Vision \& Remote Sensing, Technische Universität Berlin, Germany}}


\abstract{
We present \textit{Decomposer}, a semi-supervised reconstruction model that decomposes distorted image sequences into their fundamental building blocks - the original image and the applied augmentations, i.e., shadow, light, and occlusions. To solve this problem, we use the SIDAR dataset that provides a large number of distorted image sequences: each sequence contains images with shadows, lighting, and occlusions applied to an undistorted version. Each distortion changes the original signal in different ways, e.g., additive or multiplicative noise. We propose a transformer-based model to explicitly learn this decomposition. The sequential model uses 3D Swin-Transformers for spatio-temporal encoding and 3D U-Nets as prediction heads for individual parts of the decomposition.
We demonstrate that by separately pre-training our model on weakly supervised pseudo labels, we can steer our model to optimize for our ambiguous problem definition and learn to differentiate between the different image distortions.}

\onecolumn \maketitle \normalsize \setcounter{footnote}{0} \vfill

\setlength\parindent{0pt}
\section{INTRODUCTION}
\begin{figure*}[htbp]
    \centering
    \includegraphics[width=0.99\textwidth]{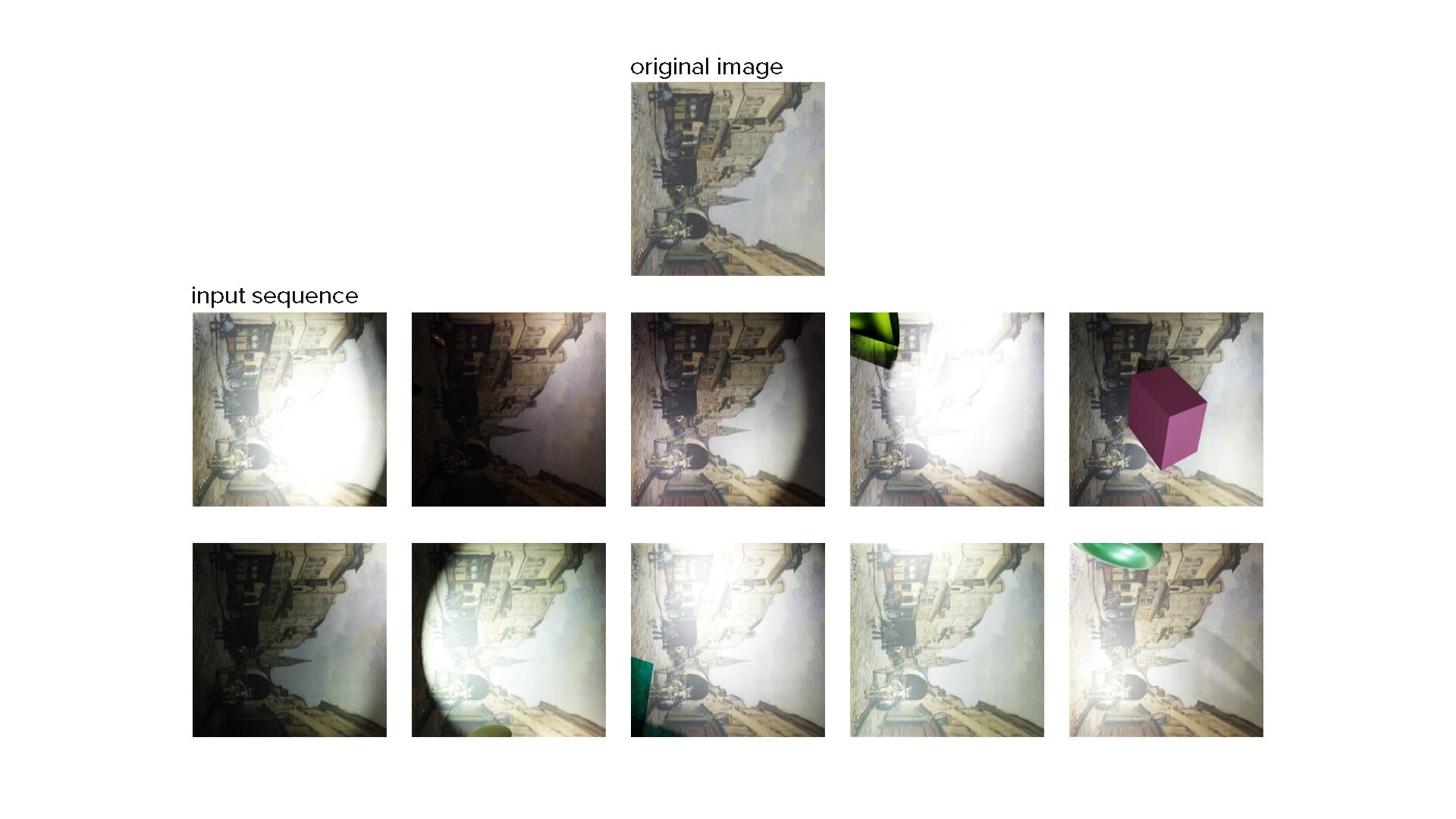}
    \caption{Example sample of SIAR dataset containing one original image and 10 augmented versions.}
    \label{fig: Example Sample}
\end{figure*}

Decomposing an image into its respective fundamental components like background, foreground, and illumination has been an important open problem in computer vision. There is a wide range of related applications such as image relighting, material recognition, image editing, object recognition and segmentation, texture transfer between images, and much more. Current approaches mostly focus on foreground detection and intrinsic image decomposition respectively, but not a combination of the two. 

We propose to use the SIDAR \cite{sidardataset} dataset that contains sequences of augmented views of an original image.
Each sequence has the same original image. 

We propose the \textbf{Decomposer} model which aims to decompose a sequence of images into their original image, element-wise applied light and shadow masks, as well as added occlusions.
Our model implements an encoder-decoder architecture to learn the latent representations of an entire sequence and extract the light, shadow, and occlusions from it.
However, the image decomposition task is an ill-posed and ambiguous problem, i.e. there can be numerous ways of decomposing an image into parts that may not correspond to the desired original image, light, shadow, and occlusions, though a combination of them still produces the input image.
For example, a model can interpret some shadows as very dark occlusions.
Therefore, we create pseudo-labels of light, shadow, and occlusions that guide our model towards a more meaningful decomposition.
Furthermore, this design choice allows our model to learn in a semi-supervised fashion.
We found that pretraining on pseudo-labels and later finetuning our model improves model performance.
\newline
\newline
Our main contributions to this image decomposition task are as follows:

\begin{itemize}
    \item We propose the \textbf{Decomposer} model, which adopts the Video SWIN Transformer as the encoder and three separate UNet models as the decoder.
    The Video SWIN Transformer, which was originally applied to videos, is applied in our model for image decomposition for sequences of images. 
    \item We create pseudo-labels for light, shadow, and occlusion masks that guide our models in learning the decomposition goals.
    \item We implement individual pre-training stages for the reconstruction of \textit{(i)} the original image, \textit{(ii)} shadow and light masks, and \textit{(iii)} occlusion masks, respectively, before finetuning the model on our dataset.
    \item We propose a mathematical definition for recombining the decomposed elements for the final reconstruction. This composition of the predicted decomposition is used for the final finetuning.
\end{itemize}

\section{RELATED WORK}
The goal of our research is related to image decomposition, where an image is separated into its formation components. Separating the occluded part from the original image is related to the topic of background subtraction. Separating the light and shadow parts from the original image is related to the topic of intrinsic images. 

\subsection{Background Subtraction}
Background subtraction is a computer vision method to detect objects in videos and separate them into background and foreground. Classical approaches include running Gaussian average, temporal median filter, mixture of Gaussians, and kernel density estimation \cite{bs}. Deep learning approaches include deep convolutional neural networks \cite{Babaee} and generative neural networks \cite{bahri}. However, these approaches make use of frame-to-frame connection in a video, which may not work well on our sequences of distorted images as there is no connection between occluded parts in an image sequence. 

\subsection{The Problem of Intrinsic Images}
The problem of intrinsic images was defined by Barrow and Tenebaum \cite{Barrow} in 1978 as recovering properties of shape, reflectance, and illumination from a single image. Since then this has been a highly challenging task in computer vision. 

Reflectance relates to the color of the material and illumination of the lighting condition. Inverse rendering estimates properties of shape, reflectance, and illumination from one or more images, where intrinsic image decomposition decomposes an image into its reflectance layer and shading layer without separating shading into shape and illumination \cite{yu}. Related works tackle this problem with classical methods, deep learning methods, as well as a combination of the two.

Classical methods perform image decomposition by returning to the physics of image formation and fitting photometric and geometric models \cite{yu}. Tappen \textit{et al.} \cite{Tappen} develops an algorithm that decomposes an image into shading and reflectance images using color and gray-scale information. Barron \textit{et al.} \cite{barron} recover intrinsic scene properties by constructing priors for shape, reflectance, and illumination and applying an optimization-based algorithm. These approaches differ from ours as they tackle the image decomposition problem by the geometry and physics in image formation, while we decompose an image based on the assumed mathematical distortions that produce it. i.e. the distorted image is produced by pixel-wise multiplication with a shadow mask and pixel-wise addition of a light mask and an occlusion mask.

Since a sophisticated physical model cannot fully capture all physical factors contributing to the appearance of a scene, deep learning approaches have been integrated into classical methods or used solely. Among them there are supervised \cite{Fan}, self-supervised \cite{Janner} and unsupervised \cite{Liu} approaches respectively. 

Yu \textit{et al.} \cite{yu} uses a fully convolutional neural network, which incorporates physical modeling of appearance approximated by local reflectance under environment illumination, to perform inverse rendering from a single, uncontrolled image. Das \textit{et al.} \cite{Das} combines physics-based priors and CNNs for intrinsic image decomposition. Zhang \textit{et al.} \cite{Zhang} design a double-stream exchange transformer network for intrinsic image decomposition, in which an independent and interrelated relationship is built between reflectance and shading, and window-based self-attention is adopted for gathering contextual information around pixels.
\section{DATASET}
To solve the problem setting of learning an image decomposition we propose to use a synthetically created dataset called SIDAR \cite{sidardataset}. It contains 15,617 sets of $1 + 10$ images, $1$ original image, and $10$ augmented versions of the original image.
An example of one sample is shown in figure \ref{fig: Example Sample}. The augmentations include shadows (lowered global ambient lighting, as well as "hard shadows"), spotlights, and occlusions in the form of shapes that occlude the original content of the image. 

\subsection{Pseudo-Labels} \label{sec:custom_labels}

The training dataset does not contain the original masks for the augmentations, i.e. it does not contain the shadow, light, and occlusion masks. Nevertheless, since we propose to "steer" our model in the right direction by pretraining our model on the individual tasks of predicting the specific augmentations, we generate approximate pseudo-labels.
In this section, we will discuss how we generated the pseudo-labels for pretraining.

\subsubsection{Light and Shadow Mask} \label{sec:sl_labels}

The shadow and light pseudo-target is the approximation of the input image without the occlusion layer.
To generate this pseudo-target, we approximate the shadow and light masks separately and multiply the original image by the shadow mask after which the light mask is added.

To generate the light mask we subtract the augmented input image from the source image and clip all values below 0. The resulting image is then blurred with a Gaussian filter to remove the remaining artifacts that come from the occlusions.
For the shadow mask, we perform the same operation, however, we first invert the augmented image.

Examples can be seen in Appendix Figure \ref{fig: SL_targets}.

\subsubsection{Occlusion Masks}

The occlusion pseudo-target approximates the binary segmentation map of occlusion and background.
An example can be seen in Appendix Figure \ref{fig: Occ_binary_targets}.

Since we generate shadow and light pseudo-targets we can pretrain our final model on those and then fine-tune the model without specific occlusion branch pertaining. We elaborate further in section \ref{sec:train_finetune}.
Using our shadow and light fine-tuned model, we can reconstruct the input images without the occlusions, only applying the predicted shadow and light masks to the original image as formalized in equation \ref{eq:sl_reconstruction}.
\begin{equation}
    \label{eq:sl_reconstruction}
    X^i_{SL} = X_{OI} \cdot X^i_{S} + X^i_{L}
\end{equation}

For each input sequence, we generate the occlusion pseudo-targets by computing the absolute difference between the augmented input image and our $X_{SL}$, apply thresholding, and convert it into a binary mask. 

\begin{equation}
    X^i_{OCC} = tresh(|X^i_{SL} - X^i|)
\end{equation}


\subsection{Evaluation Dataset}
For evaluation, on the other hand, our dataset provides a set of 100 image sequences, each including an original image with its augmented views and the occlusion mask.
Using this evaluation dataset we can quantify the performance of the individual branches of our model.
\section{MODEL}

We model an image $i$ by the composition of its fundamental building blocks, the original image $X_{OI}$, shadow $X^i_S$, light $X^i_L$, and occlusions $X^i_{OCC}$.
We first multiply the original image with the shadow mask $X^i_S$, scaling each pixel value, after which we add the light mask $X^i_L$, resulting in an image with shadow and light augmentations $X^i_{SL}$.
Finally, given the binary occlusion mask, $X^i_{M}$ we fully replace the masked parts of $X^i_{SL}$ with the occlusions in $X^i_{OCC}$.
The formula for this composition is given in equation \ref{eq:Decomposition formula}.

\begin{equation} \label{eq:Decomposition formula}
    X^i = (X_{OI} \cdot X^i_S + X^i_L) \odot X^i_{M} \cdot X^i_{OCC}
\end{equation}

\begin{figure}[h]
    \centering
    \includegraphics[width=0.5\textwidth]{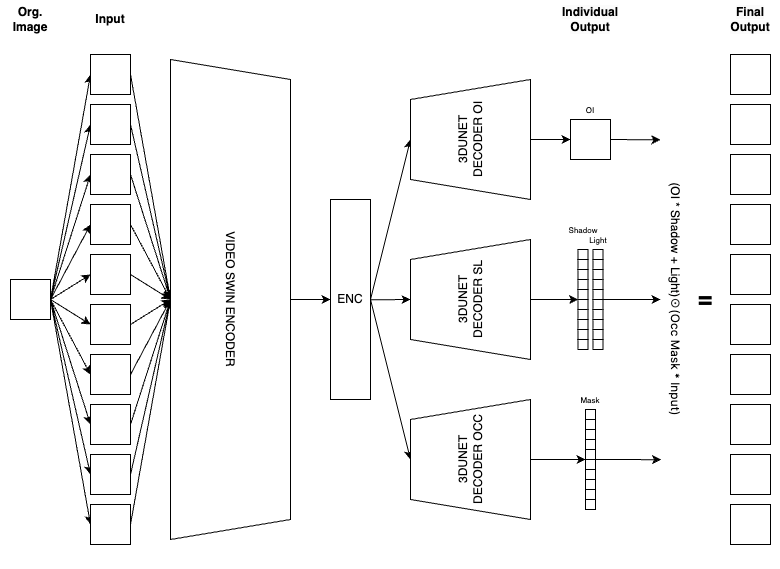}
    \caption{Decomposer model architecture. The encoder consists of a VideoSWIN network \cite{VideoSWIN} and the decoder consists of three separate 3DUnet branches with different prediction tasks.}
    \label{fig: Model Architecture}
\end{figure}

To predict the decomposition of an image we propose \textbf{Decomposer}, an encoder-decoder architecture with multiple separate decoder heads that each have the task of predicting a distinct component of the image decomposition.
Our model architecture is illustrated in figure \ref{fig: Model Architecture}.

\subsection{Encoder}
Our input data consists of a sequence of augmented versions of the same original image.
Since we are working with multiple views of the same image and want to reconstruct the entire sequence, we model this task as a sequence-to-sequence problem.
Furthermore, we want to correlate the images of our sequence in our embeddings so that the decoders can better differentiate the background and foreground in each input image.
We propose to use the VideoSWIN \cite{VideoSWIN} architecture, a sequence-to-sequence model that attends to patches across multiple images and produces new embeddings for each individual image whilst merging information of the entire sequence.
The final embedding dimensions are $CxSxHxW$, where $C$ is the number of channels, $S$ is the sequence length, and $H$ and $W$ are height and width, respectively.

\subsection{Decoders}
To operate on 3D embeddings we implement the 3DUNet \cite{3DUNet} architecture as our decoder.
We use three separate decoders, one for predicting the single original image, one for predicting the shadow and light masks, and one for predicting the binary occlusion mask.
We have deliberately selected three separate decoders because the features extracted for generating the original image are different from those for predicting shadow and light, and occlusions, respectively.

Our original image decoder branch predicts a single RGB image. This image is the unaugmented original image.

Our shadow and light decoder branch predicts two one-dimensional masks that correspond to the shadow and light masks for each input image.

Our occlusion branch predicts a one-dimensional output for the binary occlusion mask for each input image.

The final output consists of the composition of the predicted elements following equation \ref{eq:Decomposition formula} for each individual input image $X^i$.
\section{TRAINING}
We found that training the model in multiple stages resulted in the best results. Training was split into 5 stages which will be discussed in the following sections. Section \ref{sec:train_pretrain} discusses the pretraining of the encoder. This subsequently enables pretraining the decoders (discussed in Sections \ref{sec:train_gt}, \ref{sec:train_sl} and \ref{sec:train_occ}). Finally, all decoder heads are trained together in a fine-tuning stage which includes all decoder heads and follows Equation \ref{eq:Decomposition formula} for its objective. This is discussed in Section \ref{sec:train_finetune}.

When pretraining the individual decoder heads, those that are not to be trained are frozen.

\subsection{Encoder Pretraining} \label{sec:train_pretrain}
To stabilize further fine-tuning and to ensure the encoding retains all necessary information to recreate the whole input-sequence a pretraining for the SWIN encoder was performed in an auto-encoder fashion. This was done using one UNet-decoder outputting 10 images.

The auto-encoder was trained for 300 epochs, with the validation loss steadily decreasing. After 300 epochs the training was stopped due to time restraints and satisfactory results, and only the SWIN-encoder was saved which reconstructed all the inputs in satisfactory detail, although having decreased sharpness/clarity.
This was likely due to the depth of the SWIN encoder and the representation being too compressed.
We deemed the results satisfactory as they suffice for the task.
In all further experiments, this checkpoint was used and the experiment set the "benchmark" for sharpness.

\subsection{OI Branch Pretraining} \label{sec:train_gt}
The first UNet decoder head to be pretrained was the Original Image ($OI$) branch.
This was done as it served as a base for the other heads in later pretraining.
From the SWIN encoder pretraining it was known that the resulting OI reconstruction would decrease in sharpness compared to the original which was deemed satisfactory. 

The frozen weights of the OI UNet head can be used as a base on which shadow, light, and occlusion were added.
Nevertheless, for training other branches we used the ground truth original image instead of the predicted one to avoid the reduced sharpness which potentially introduced noise.

\subsection{Shadow and Light Branch Pretraining} \label{sec:train_sl}
Given approximations of the shadow and light ($SL$) masks (see Section \ref{sec:sl_labels}), our $SL$-branch can be pretrained where the shadow and light view of the input $X^i_{SL}$ is constructed from the ground truth original image and the models $SL$ prediction as formalized in Equation \ref{eq:sl_reconstruction}.

The goal of this pretraining is to direct the output "in the right direction" for the weights to be in the right "area in the loss landscape" and for the model to approximately know what shadow and light are.

As the approximated $SL$ masks do not contain the hard shadows and spotlights in the original data, the pretrained branch does not predict these and therefore needs further fine-tuning at a later stage.


\subsection{Occlusion Branch Pretraining} \label{sec:train_occ}
Like in the Shadow and Light pretraining, approximated targets were generated to steer the Occlusion (OCC) UNet decoder in the correct direction.
Binary Cross Entropy Loss with weighted classes was used as the class distribution is heavily biased towards the background class.

\subsection{Fine-Tuning} \label{sec:train_finetune}
In the pretraining stages approximated targets were used, which steer the decoder heads in the correct direction but do not yield perfect results.
To further improve results to better resemble the input augmented images, all decoder heads, i.e. the model as a whole with frozen encoder, were fine-tuned using a weighted sum of the Mean Absolute Error between each input image and the respective predicted composition following Equation \ref{eq:Loss decomp} and the Mean Absolute Error between the predicted $OI$ reconstruction and the ground truth $OI$ (Equation \ref{eq:Loss oi}).

\begin{align}
    \mathcal{L}_{oi} &= | \hat{X}_{OI} - X_{OI} | \label{eq:Loss oi} \\
    \mathcal{L}_{decomp} &= \sum^{10}_{i=1}  | X^i - (X_{OI} \cdot X^i_S + X^i_L) \odot X^i_{M} \cdot X^i_{OCC} |  \label{eq:Loss decomp}
\end{align}

We propose a further regularization, mask decay, that prevents our occlusion branch from predicting too many false positives.
Mask decay ($md$) is inspired by weight decay and implements the weighted mean of the predicted binary mask, which in the final loss is subject to minimization.
This forces the model to only predict a positive value when there is an actual occlusion in the input image.

The final loss is defined in Equation \ref{eq:Loss}.
\begin{equation}
    \mathcal{L} = \alpha_{decomp} \mathcal{L}_{decomp} + \alpha_{oi} \mathcal{L}_{oi} + md \label{eq:Loss}
\end{equation}



Given that all decoder branches are pretrained and vaguely return the desired output, fine-tuning for 600 epochs was performed.
Fine-tuning after pretraining and regularization through the loss $\mathcal{L}$ improves the individual decoder predictions and the final reconstruction.

\section{RESULTS}

In this section, we assess the performance of our model on the decomposition task.
Firstly, we perform a quantitative evaluation of the individual outputs of our (i) $OI$- and (ii) $SL$-branches, and (iii) of our final predicted reconstruction.
We then perform a quantitative evaluation of all individually predicted images and the final reconstruction.
Moreover, we discuss the shortcomings of our model.

\subsection{Quantitative} 

We employ two metrics to compute the similarity between images. Firstly, we propose to use the SSIM metric which assesses similarity in texture and patterns between two images.
Furthermore, we use the mean squared error (MSE) for relative comparisons, which is more content-based in nature. 
An SSIM score equal to 1.0 indicates identical images, while a value of 0.0 suggests completely different images. Additionally, a higher MSE shows a more significant difference between the two images.

To calculate the quantitative results, we evaluate the similarity between our original image reconstruction and our predicted light and shadow reconstruction $X^i_{SL}$ to their respective ground truth targets.

\subsubsection{OI Reconstruction} \label{sec:oi_quant}
Table \ref{quantitative_results} presents the results of our quantitative experiments on the OI reconstruction task.
The original image reconstruction has an average SSIM score of 0.67, signifying a high texture and structure similarity between the ground truth and its reconstruction.
Additionally, we achieved an average MSE of 1102.
To provide a comparison, for each original image in the evaluation dataset, we selected a random ground truth image and calculated the MSE and SSIM between them.
Randomly chosen images had an average MSE of 9373 and SSIM of 0.14.
Thus we assume that our reconstruction performs significantly better than the random experiment.
Consequently, we conclude that the reconstructed images are similar in texture, structure, and content to their ground truth targets. 

\begin{table}[!h]
    \centering
    \begin{tabular}{|c| c | c|} 
         \hline
         Experiments & MSE & SSIM \\ [0.5ex] 
         \hline
         OI reconstruction & 1102 & 0.67\\ 
         \hline
         Random OI & 9373 & 0.14 \\
         \hline
    \end{tabular}
    \caption{Results of the original image reconstruction (OI). In OI reconstruction we compare the original image to our reconstructed result. Random OI comparing two random (different) ground truth images.}
    \label{quantitative_results}
\end{table}

\subsubsection{SL and Full Reconstruction} \label{sec:quant_sl_and_full}
We construct our predicted shadow-and-light augmented image $X^i_{SL}$ following equation \ref{eq:sl_reconstruction} to evaluate shadow and light masks.
Unfortunately, we do not have access to images without occlusions.
Therefore, to remove the influence of occlusions in our evaluation, we mask the regions with occlusions using the occlusion ground truth masks in both $X^i_{SL}$ and the input images $X^i$ that function as targets.

The results of that experiment are shown in the first row of table \ref{quantitative_results_shadow_light}.
We obtained a very high SSIM score of 0.82 and a very low MSE averaged over each sequence.
This indicates that the model performs well in predicting shadow and light components.

The composition of all predicted elements of the image decomposition is our final, full reconstruction.
We achieve an SSIM score of 0.88, which is an improvement over 0.82 when evaluating solely the shadow and light views $X^i_{SL}$.
This indicates that the predicted masks improve the overall structural similarity between predicted reconstruction and input.

\begin{table}[!h]
    \centering
    \begin{tabular}{|c| c | c|} 
        \hline
        Experiments & MSE & SSIM \\ [0.5ex] 
        \hline
        Shadow/light reconstruction & 268 & 0.82\\ 
        \hline
        Full reconstruction & 220.5 & 0.88\\
        \hline

    \end{tabular}
    \caption{Results of the shadow and light reconstruction (with masked occlusions) and full input image reconstruction with shadow, light and occlusions.}
    \label{quantitative_results_shadow_light}
\end{table}

One significant consideration is the fact, that predicted occlusion masks select which parts of the input should we use for the final predicted reconstruction.
This means, that if our masks predict too many positive values, in the extreme a completely positive mask, the input is compared to itself, yielding a perfect SSIM and MAE score.
Nevertheless, due to our mask-decay regularization and proper pretraining, we force the occlusion branch to significantly reduce the false positive rate and achieve one of $4.4\%$ during our evaluation.

In further experiments, we observe that removing the $OI$ sub-loss ($\mathcal{L}_{oi}$) in our final Loss (Equation \ref{eq:Loss}) reduces the quality of the $OI$ predictions as the prediction significantly oversaturates.
Nevertheless, this improves the overall performance of the final $SL$ reconstruction from an SSIM score of 0.82 to 0.91, thus also improving the final reconstruction.
An emerging property of the predicted shadow and light masks is a color-correcting effect that desaturates and sharpens the final reconstruction.

Nevertheless, we decided to continue with a higher quality $OI$ prediction, as the focus of the research is on the individual branch predictions, rather than a perfect final reconstruction.

\subsection{Qualitative} 
For the qualitative results, we look separately at every component of the decomposition and (if available) we compare it with ground truth.

\begin{figure}[h]
    \centering
    \begin{subfigure}[a]{0.45\textwidth}
       \includegraphics[width=1\linewidth]{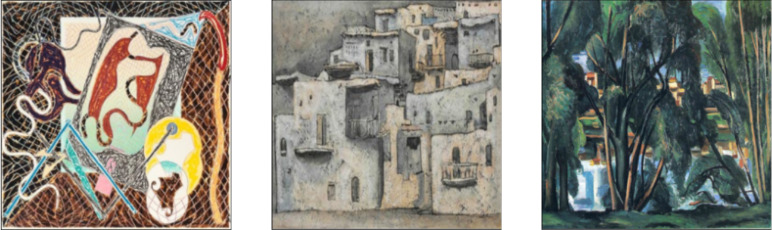}
       \caption{Input}
       \label{fig:results_input_gt} 
    \end{subfigure}

    \begin{subfigure}[b]{0.45\textwidth}
       \includegraphics[width=1\linewidth]{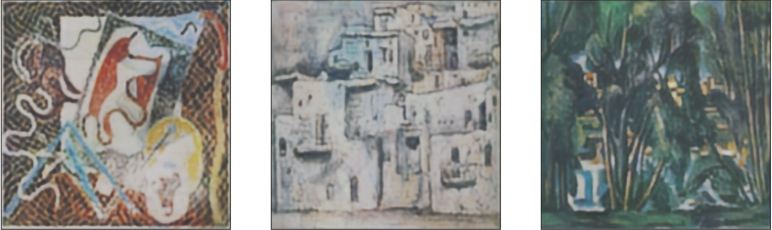}
       \caption{Output}
       \label{fig:results_output_gt} 
    \end{subfigure}

    \caption{Original image reconstruction of the final model}
    \label{fig:result_oi_recon}
\end{figure}

\begin{figure*}[htbp]
    \centering
    \includegraphics[width=1.0\linewidth]{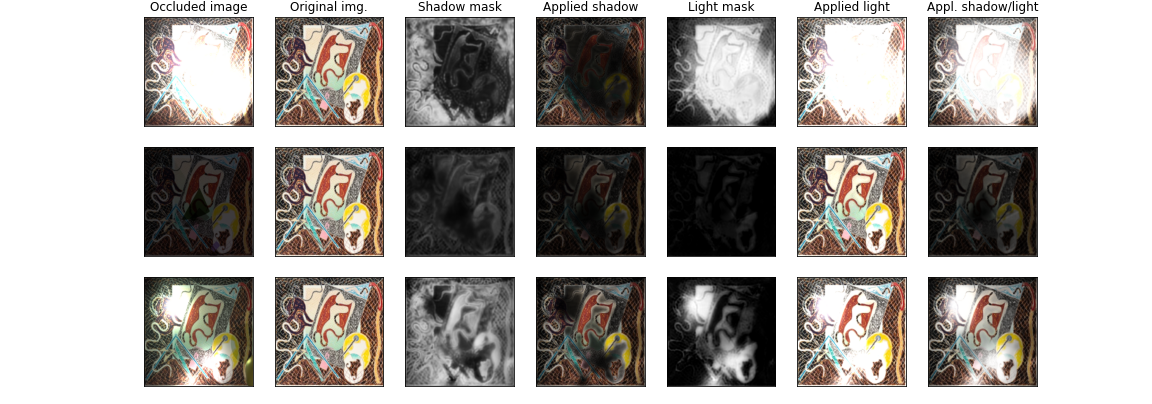}
    \caption{Results of shadow and light decomposition. The left columns shows the input sequence and estimated the original image. The estimated shadow and light masks are visualized and applied to the original image. The last column shows the composition of all artifacts.}
    \label{fig:result_shadow_light}
\end{figure*}

\subsubsection{Original Image Reconstruction}
Figure \ref{fig:result_oi_recon} shows the original image reconstruction results.
Evidently, the model is able to reconstruct fine textures from the input image.
Additionally, we have successfully removed most shadow, light, and occlusion artifacts from the image.
However, as discussed in Section \ref{sec:oi_quant}, the model seems to undersaturate colors and reduce opacity when compared to the ground truth original image.
Nevertheless, when combined with the shadow and light mask, the final reconstruction yields very promising results.

\subsubsection{Shadow and Light}
Figure \ref{fig:result_shadow_light} shows predicted and applied shadow and light masks.

Dark areas in the shadow mask, i.e. predicted values close to 0 influence the reconstruction the most since the shadow mask is a multiplicative term.
Column 3 of Figure \ref{fig:result_shadow_light} shows the results of the shadow decomposition.
The model properly predicts the global ambient shadow and the harsh shadows.
Counterintuitively, the shadow mask is 0 in areas where the input image has strong light thus introducing strong shadows in those regions.
This is due to the balancing act the model has learned to perform when combining shadow and light in the final composition of our predicted individual shadow and light masks.

In contrast to the shadow mask, the light mask is an additive term and affects the composition inversely to the shadow mask.
Bright areas, i.e. predicted values close to 1 influence the reconstruction the most.
In column 5 of Figure \ref{fig:result_shadow_light} we observe, that the light mask reconstructs all strong light sources from the input image.
By applying the light mask to the original image, we effectively recreate the lighting conditions.
Finally, as demonstrated in column 7, adding the predicted shadows enables us to reconstruct the complete shadow-and-light augmented image without occlusions.

Very dark occlusions turned out to be a significant challenge for the model.
This difficulty comes from the complexity of distinguishing between harsh shadows and intensely dark occlusions.
Thus, dark occlusions are very often inaccurately reconstructed as shadows, as illustrated in row 3 of Figure \ref{fig:result_shadow_light}.
A similar effect occurs when dealing with bright (white) occlusions and light decomposition.

\begin{figure}
    \centering
    \includegraphics[width=0.8\linewidth]{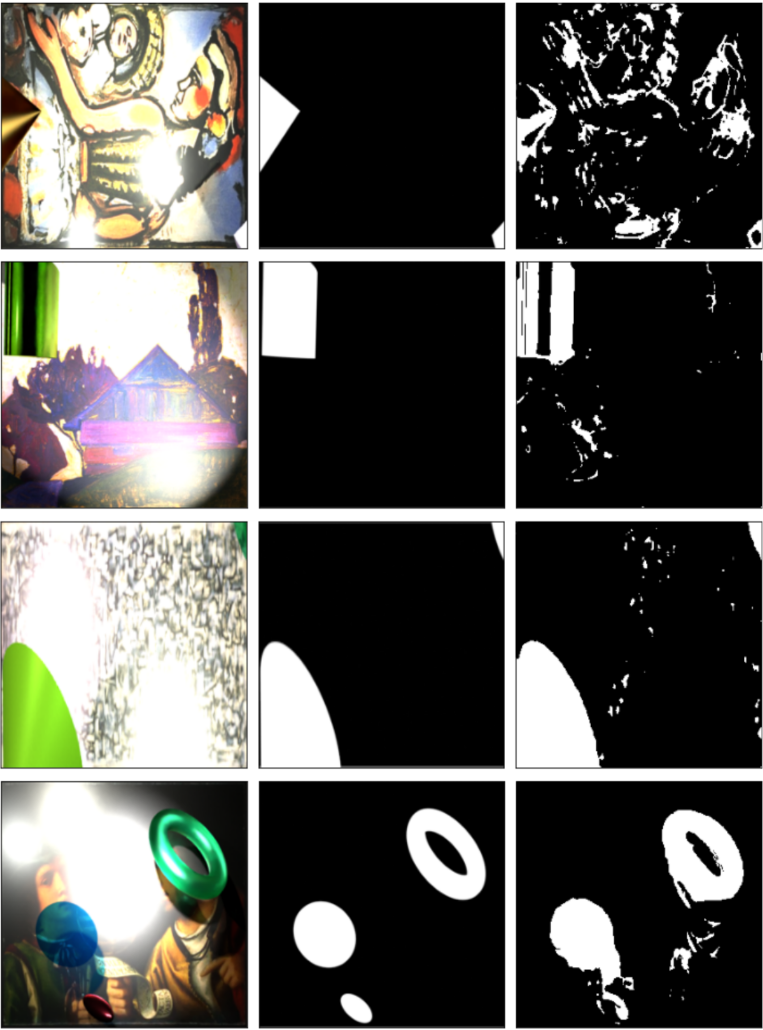}
    \caption{Occlusion binary masks after fine-tuning. The left column shows the original input image, the middle column the ground truth binary occlusion mask, and the right column the predicted binary occlusion mask.}
    \label{fig:result_occlusion}
\end{figure}

\subsubsection{Occlusions}
Figure \ref{fig:result_occlusion} demonstrates the results of predicted occlusion masks.
We observe that our predicted masks successfully capture occlusions that strongly separate themselves from the background, e.g. in row 3.
Moreover, even if the occlusions are transparent, as in row 4, our model properly segments the occlusions.
Nevertheless, due to the ambiguity of very dark occlusions, which could also be interpreted as shadows, the model fails to interpret darker parts of occlusions as occlusions as seen in row 2.
This ambiguity extends to very bright spotlights, as those obscure the original image in the same way an occlusion would.
We can observe this effect in row 1.

\subsubsection{Full Reconstruction}
\begin{figure}
    \centering
    \includegraphics[width=1\linewidth]{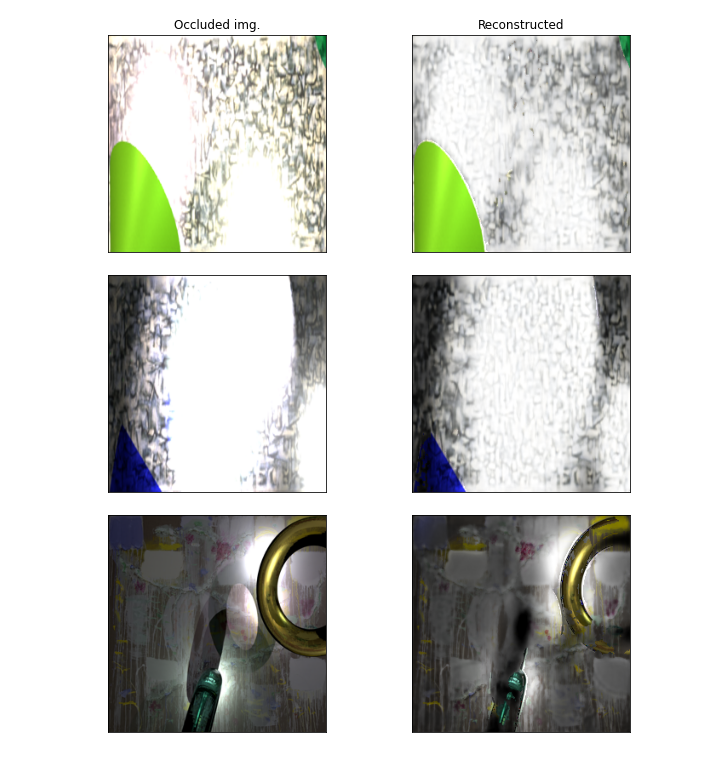}
    \caption{Input image reconstruction using shadow, light, and occlusion masks.}
    \label{fig:full_reconstruction}
\end{figure}

Finally, the model's objective subject to optimization was to reconstruct augmented images by applying formula \ref{eq:Decomposition formula}.
Figure \ref{fig:full_reconstruction} shows the result of composing all predicted elements of the decomposition to reconstruct the input.
We observe the discussed effects of applying shadow and light and can now additionally see the integration of masked occlusions.
Notably, the predicted occlusion masks are slightly larger than the actual occlusions, resulting in visible borders.
Nevertheless, as discussed in Section \ref{sec:quant_sl_and_full}, we achieve a small false positive rate of $4.4\%$.


\section{CONCLUSION}

In this work, we propose \textbf{Decomposer}, a single model capable of decomposing input images of an image sequence into their respective core components: the underlying original image, applied shadow and light, and added occlusions.
We demonstrate that using automatically created pseudo-labels in combination with pretraining the individual branches of a sequence-to-sequence encoder-decoder model can guide the distinct prediction heads towards the desired directions, which can then be further improved by fine-tuning jointly.










\bibliographystyle{apalike}
{\small
\bibliography{example}}

\begin{thebibliography}{}

\bibitem[Babaee et~al., 2017]{Babaee}
Babaee, M., Dinh, D.~T., and Rigoll, G. (2017).
\newblock A deep convolutional neural network for background subtraction.
\newblock {\em CoRR}, abs/1702.01731.

\bibitem[Bahri and Ray, 2022]{bahri}
Bahri, F. and Ray, N. (2022).
\newblock Dynamic background subtraction by generative neural networks.

\bibitem[Barron and Malik, 2020]{barron}
Barron, J.~T. and Malik, J. (2020).
\newblock Shape, illumination, and reflectance from shading.

\bibitem[Barrow and Tenenbaum, 1978]{Barrow}
Barrow, H.~G. and Tenenbaum, J.~M. (1978).
\newblock Recovering intrinsic scene characteristics from images.

\bibitem[Das et~al., 2022]{Das}
Das, P., Karaoglu, S., and Gevers, T. (2022).
\newblock Intrinsic image decomposition using physics-based cues and cnns.
\newblock {\em Computer Vision and Image Understanding}, 223:103538.

\bibitem[Fan et~al., 2018]{Fan}
Fan, Q., Yang, J., Hua, G., Chen, B., and Wipf, D. (2018).
\newblock Revisiting deep intrinsic image decompositions.

\bibitem[Janner et~al., 2017]{Janner}
Janner, M., Wu, J., Kulkarni, T.~D., Yildirim, I., and Tenenbaum, J.~B. (2017).
\newblock Self-supervised intrinsic image decomposition.
\newblock {\em CoRR}, abs/1711.03678.

\bibitem[Kwiatkowski et~al., 2023]{sidardataset}
Kwiatkowski, M., Matern, S., and Hellwich, O. (2023).
\newblock Sidar: Synthetic image dataset for alignment \& restoration.

\bibitem[Liu et~al., 2020]{Liu}
Liu, Y., Li, Y., You, S., and Lu, F. (2020).
\newblock Unsupervised learning for intrinsic image decomposition from a single image.
\newblock In {\em 2020 IEEE/CVF Conference on Computer Vision and Pattern Recognition (CVPR)}, pages 3245--3254.

\bibitem[Liu et~al., 2021]{VideoSWIN}
Liu, Z., Ning, J., Cao, Y., Wei, Y., Zhang, Z., Lin, S., and Hu, H. (2021).
\newblock Video swin transformer.

\bibitem[Piccardi, 2004]{bs}
Piccardi, M. (2004).
\newblock Background subtraction techniques: a review.
\newblock In {\em 2004 IEEE International Conference on Systems, Man and Cybernetics (IEEE Cat. No.04CH37583)}, volume~4, pages 3099--3104 vol.4.

\bibitem[Tappen et~al., 2005]{Tappen}
Tappen, M., Freeman, W., and Adelson, E. (2005).
\newblock Recovering intrinsic images from a single image.
\newblock {\em IEEE Transactions on Pattern Analysis and Machine Intelligence}, 27(9):1459--1472.

\bibitem[Yu and Smith, 2018]{yu}
Yu, Y. and Smith, W. A.~P. (2018).
\newblock Inverserendernet: Learning single image inverse rendering.

\bibitem[Zhang et~al., 2022]{Zhang}
Zhang, F., Jiang, X., Xia, Z., Gabbouj, M., Peng, J., and Feng, X. (2022).
\newblock A double-stream exchange transformer network for intrinsic image decomposition.
\newblock In {\em 2022 International Conference on Image Processing and Media Computing (ICIPMC)}, pages 51--55.

\bibitem[Özgün Çiçek et~al., 2016]{3DUNet}
Özgün Çiçek, Abdulkadir, A., Lienkamp, S.~S., Brox, T., and Ronneberger, O. (2016).
\newblock 3d u-net: Learning dense volumetric segmentation from sparse annotation.

\end{thebibliography}

\appendix

\section*{\uppercase{Appendix}} \label{Appendix}


        

\begin{figure}[h]
\begin{center}
    \begin{minipage}{0.45\textwidth}
        \centering
        \includegraphics[width=1.0\textwidth]{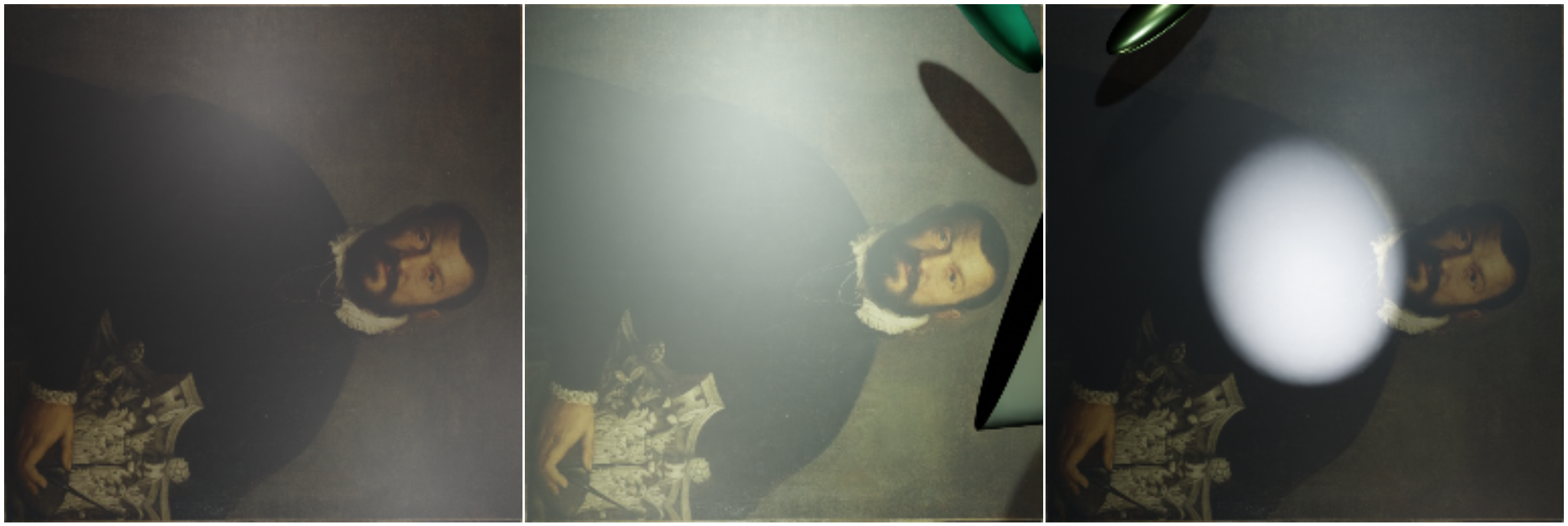} 
    \end{minipage}\hfill
    \begin{minipage}{0.45\textwidth}
        \includegraphics[width=1.0\textwidth]{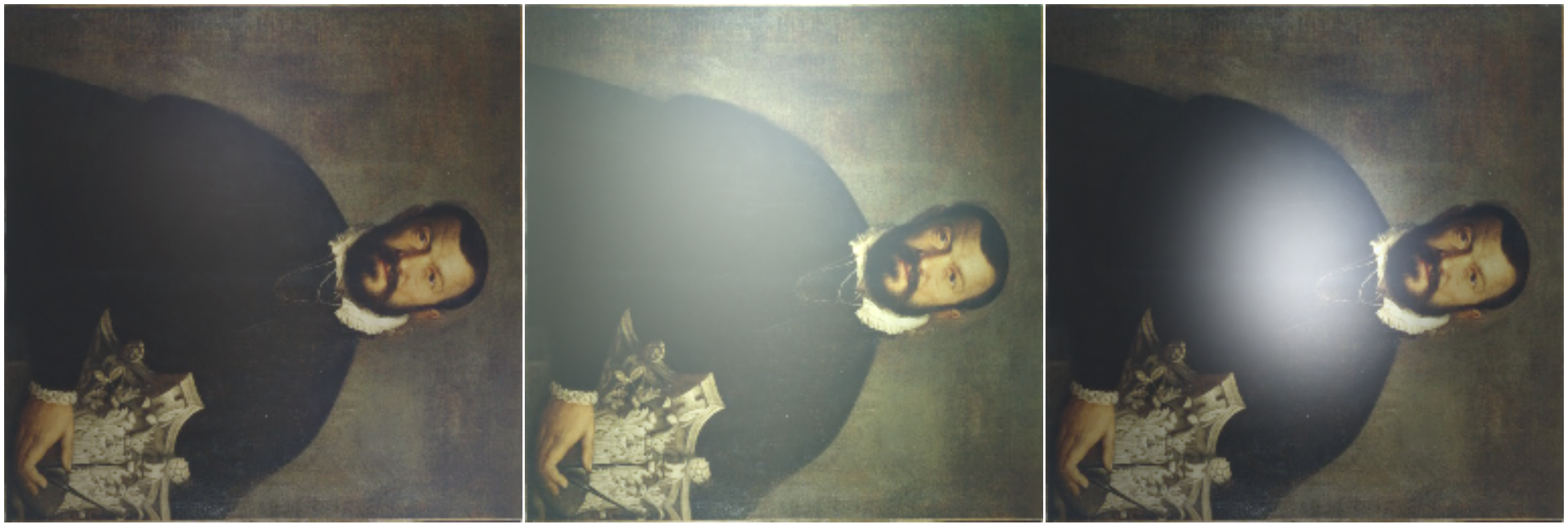} 
    \end{minipage}
\end{center}
\caption{Example for generated shadow and light pseudo-target. Top: Input Image. Bottom: Pseudo target.}
\label{fig: SL_targets}
\end{figure}

\begin{figure}[h]
\begin{center}
    \begin{minipage}{0.45\textwidth}
        \centering
        \includegraphics[width=1.0\textwidth]{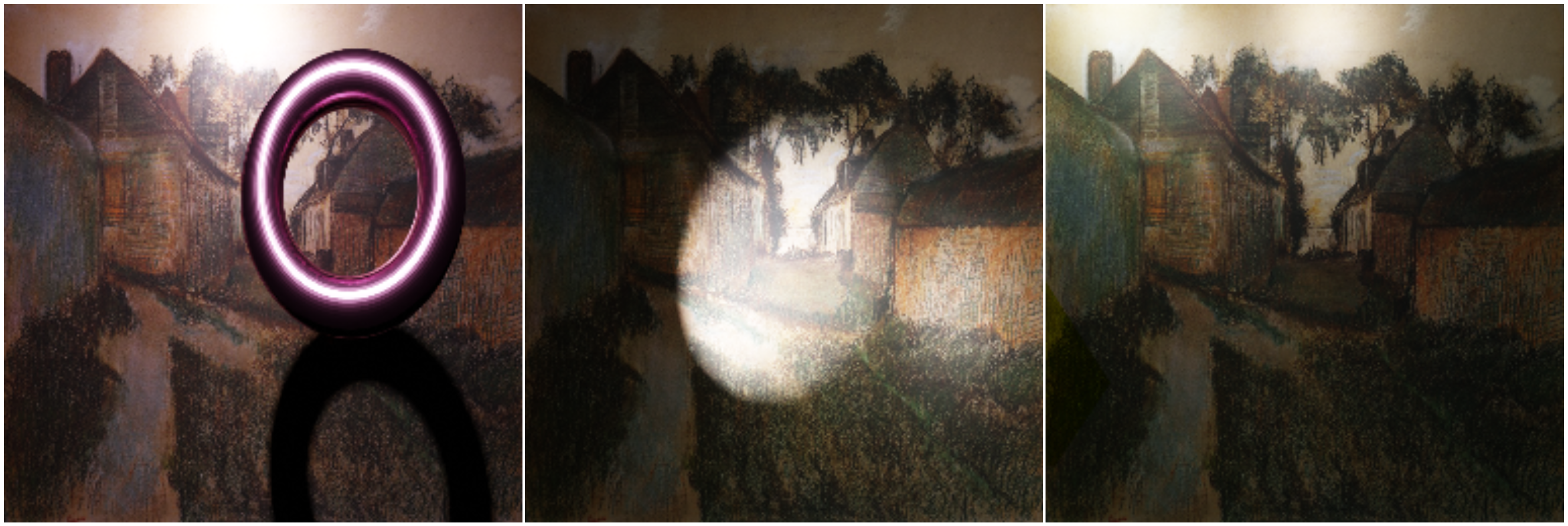} 
    \end{minipage}\hfill
    \begin{minipage}{0.45\textwidth}
        \includegraphics[width=1.0\textwidth]{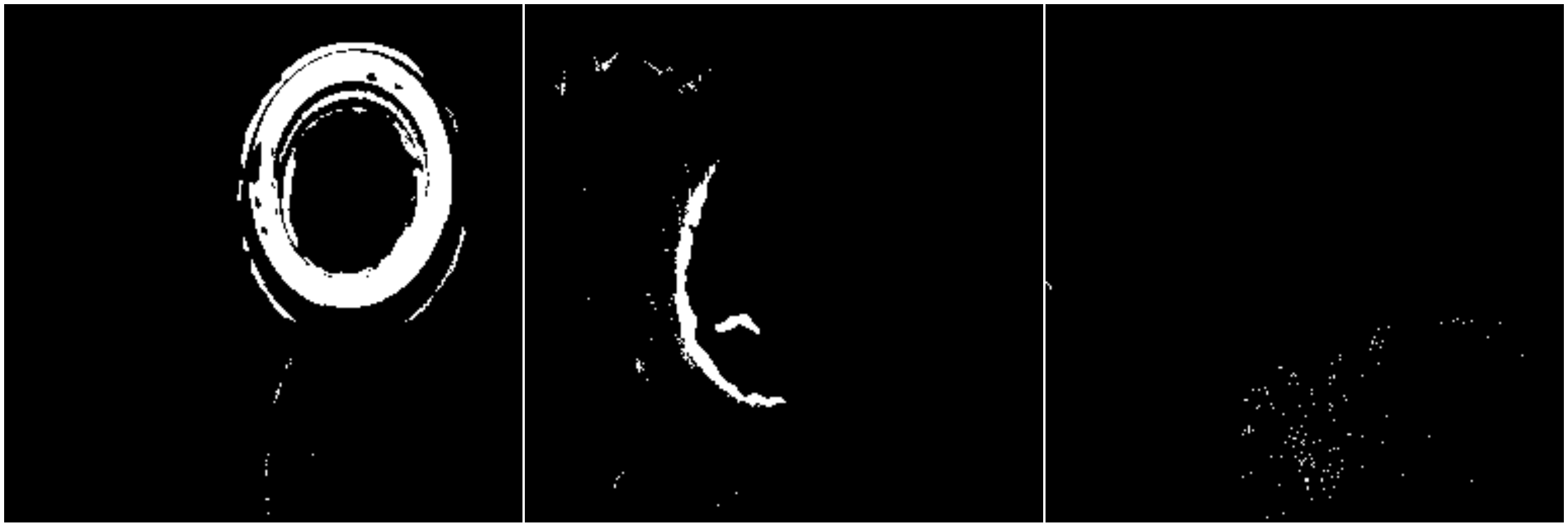} 
    \end{minipage}
\end{center}
\caption{Example for generated occlusion mask pseudo-target. Top: Input Image. Bottom: Pseudo target.}
\label{fig: Occ_binary_targets}
\end{figure}

\end{document}